\documentclass{article}
\usepackage[preprint]{spconf}

\copyrightnotice{\footnotesize
  \begin{minipage}{\textwidth}
  \centering
  \textcolor{red}{Copyright~\copyright~2018 IEEE. Personal use of this material is permitted. Permission from IEEE must be obtained for all
other uses, in any current or future media, including reprinting/republishing this material for advertising
or promotional purposes, creating new collective works, for resale or redistribution to servers or lists, or
reuse of any copyrighted component of this work in other works. }
  \end{minipage}}
\usepackage{amsmath}

\usepackage{graphicx}
\usepackage{amsmath}
\usepackage{amsthm}
\usepackage{amssymb}

\usepackage{algorithm} 
\usepackage{algpseudocode}
\usepackage{framed}
\usepackage{xcolor}
\usepackage{varwidth}
\definecolor{shadecolor}{gray}{0.95}
\usepackage{enumitem}

\newtheorem{definition}{Definition}[section]

\title{Peekaboo $-$ Where are the Objects? \\ Structure Adjusting Superpixels}
\makeatletter
\def\@name{\emph{Georg Maierhofer$^{1\ast}$, Daniel Heydecker$^{1\ast}$, Angelica I. Aviles-Rivero$^{1\ast}$,} \thanks{* These three authors contributed equally and hold joint first authorship. }\thanks{This work was supported by the UK Engineering and Physical Sciences Research Council (EPSRC) grant EP/L016516/1 for the University of Cambridge Centre for Doctoral Training, the Cambridge Centre for Analysis. Support from the CMIH University of Cambridge is greatly acknowledged.}\\ \emph{Samar M. Alsaleh$^{2}$ and Carola-Bibiane Sch{\"o}nlieb}$^{1}$}
\address{$^{1}$DAMTP and DPMMS, Faculty of Mathematics, University of Cambridge, UK.\\
$^{2}$ Department of Computer Science, George Washington University, USA.\\
}
\def\ps@preprint{
  \def\@oddfoot{\mycopyrightnotice}
  \def\@evenfoot{}
}
\def\mycopyrightnotice{
  {\footnotesize
  \begin{minipage}{\textwidth}
  \centering
  Copyright~\copyright~2017 IEEE. Personal use of this material is permitted. However, permission to use this  \\
  material for any other purposes must be obtained from the IEEE by sending a request to pubs-permissions@ieee.org.
  \end{minipage}
  }
}

\begin{document}
%
\maketitle

%
%
%
%
\begin{abstract}
This paper addresses the search for a fast and meaningful image segmentation in the context of $k$-means clustering. The proposed method builds on a widely-used local version of Lloyd's algorithm, called Simple Linear Iterative Clustering (SLIC). We propose an algorithm which extends SLIC to dynamically adjust the local search,  adopting superpixel resolution dynamically to structure existent in the image, and thus provides for more meaningful superpixels in the same linear runtime as standard SLIC. The proposed method is evaluated against state-of-the-art techniques and improved boundary adherence and undersegmentation error are observed, whilst still remaining among the fastest algorithms which are tested.
\end{abstract}
\begin{keywords}
Image segmentation, Clustering algorithms, Image texture analysis
\end{keywords}
\section{Introduction}
\label{sec:intro}

Image segmentation continues to be a focus of great attention in the field of computer vision. This is, primarily, because segmentation is a key pre-processing step in a broad range of applications (e.g.~\cite{ngo2015automatic,huang2016stem,shen2017efficient}). In particular, superpixel segmentation is a prominent technique that has been applied to a wide range of computer vision tasks including object detection~\cite{yan2015object}, depth estimation~\cite{liu2015deep}, optical flow~\cite{lu2013patch,menze2015object} and object tracking~\cite{wang2011superpixel}. The idea of superpixels is to divide the image into multiple clusters, which ideally reflect qualities such as similar colour, and boundaries overlapping with existing boundaries in the image.
%
%
%

Superpixel segmentation as a stand-alone tool decreases computational load by reducing the number of primitives in the image domain, while at the same time increasing discriminative information~\cite{SLIC::2012,stutz2017superpixels}.  These factors have motivated the fast development of diverse superpixel segmentation techniques starting from the pioneering work of Ren and Malik~\cite{ren2003learning} and followed by diverse approaches including the ones reported in~\cite{SLIC::2012,vedaldi2008quick,van2012seeds,tang2012topology}.
%
%
%
%
\begin{figure}[!h]
\centering
\includegraphics[width=0.5\textwidth]{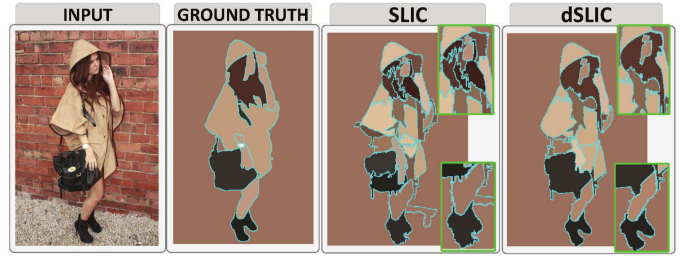}
\caption{Visual comparison of SLIC vs our dSLIC. (From left to right) Input image, ground truth and outputs from SLIC and dSLIC.
Zoom-in views show output details, in which dSLIC avoids segmenting large uniform domains (e.g. hair and bag).}
\label{fig::introFigB}
\end{figure}

In particular, the Simple Linear Iterative Clustering (SLIC) \cite{SLIC::2012} algorithm  is a top reference solution, and probably the most widely-used approach, that fulfils desirable properties such as computational tractability and good boundary adherence. SLIC performs image segmentation into superpixels, building on Lloyd's algorithm~\cite{lloyd1982least} for $k$-means clustering. The general approach is an iterative scheme which adapts the segmentation at each iteration such that the updated domains consist of points which are similar, in distance or colour, to a particular segment in the previous step.
%
%
%
%

The central observation of SLIC is that using a similarity-metric, which includes spatial distance in the image, one can justify a limitation on the search range for the updates. This limit yields to a vastly improved computational performance of SLIC when compared to traditional segmentation algorithms. However, the introduction of a limited search range also has downsides: (1) large uniform domains are segmented into unnecessarily small superpixels and (2) in regions with more structure, i.e. more objects, the final superpixel size is much smaller than the search radius of SLIC; thus, many distance calculations are actually obsolete and computational efficiency could be further enhanced.

Motivated by the aforementioned drawbacks, in this work we propose an extension of the SLIC algorithm which we call dSLIC, where d stands for dynamic.  Our solution allows for a \emph{dynamic search field} in each iteration, thus allowing the algorithm and final superpixels to better adjust to structure which is prevalent in the image (see Fig.~\ref{fig::introFigB}). While this is an important part of our solution \textbf{our contributions are:} (1) We propose a structure measure that allows for dynamic adjustment of the search range according to the density of objects/boundaries. (2) We show that our proposed distance calculations allows: (i) searching, in structure-poor parts, larger domains to connect meaningful uniform regions and (ii) reducing, in structure-rich parts, the search radius to save computational resources. (3) We demonstrate that injecting our structure measure to the SLIC computation leads to a segmentation closer to the ground-truth. (4) We provide evidence of the general applicability  of our solution with several datasets, and compare against some work from the body of literature.


\section{Background $-$ From Lloyd to SLIC}

In this section, we discuss the basis of our dSLIC solution, based on the connection between the k-means clustering described in~\cite{lloyd1982least} and the SLIC algorithm~\cite{SLIC::2012}.

%
%

Consider an input image of integer width $w$ and integer height $h$ as $I:[w]\times[h]\rightarrow D$, where $D$ is the image domain.
For our purpose, $D$ could be $[0,1]$ for greyscale images or a subset of $\mathbb{R}^3$ for colour images. Then, segmenting the input $I$ into superpixels based on $k$-means clustering  can be formulated as a minimisation problem in the following form:

\begin{definition}[Segmentation in $k$-means clustering]
Given an image $I:\mathcal{X}\rightarrow D$, where $\mathcal{X}=[w]\times[h]\subset \mathbb{Z}^2$, a segmentation into superpixels is a partition $\{S_i\}_{i=1}^n$ of $\mathcal{X}$ such that:

\begin{enumerate}[leftmargin=*,noitemsep]
    \item Each $S_i$ is path-connected, with respect to the usual grid on $\mathbb{Z}^2$.
    \item For each $1\leq i\leq n$ we have $S_i=\{x:d((x,I(x)),F(S_i))\\=\min_{1\leq j\leq n} d((x,I(x)),F(S_j))\}$.
\end{enumerate}

\noindent
where $d$ is a metric on the space $\mathcal{X}\times D$ and $F:\mathcal{P}(\mathcal{X})\rightarrow\mathcal{X}\times D$ is the feature function on the set of all partitions of the image grid.
\end{definition}

\noindent
Intuitively speaking, $F$ represents the main features of each cluster and $d$ is a measure for similarity between any two points in the space $[w]\times[h]\times D$.
%
%

A fairly traditional approach to image segmentation is $k$-means clustering and in particular Lloyd's algorithm~\cite{lloyd1982least}. The idea is to compute the point in $\mathcal{X}$ closest to the mean (which we shall call the cluster center) of all cluster points, as $F(S):=\left[\frac{1}{|S|}\sum_{x\in S}(x,I(x))\right]$, and to update the segmentation iteratively, ensuring at each step that we assign points to the nearest cluster from the previous step. Inspired by this concept, Achata et al. in~\cite{SLIC::2012} proposed the SLIC algorithm which is a local version of the Lloyd's algorithm.

In the setting of SLIC, a distance measure is used as:  For $\mathbf{p}_1,\mathbf{p}_{2}\in \mathcal{X}\times D, \mathbf{p}_i=[\mathbf{x}_i,\mathbf{l}_i]^T,\mathbf{x}_i\in\mathcal{X},\mathbf{l}_i\in D$ define
\begin{align}
    d(\mathbf{p}_1,\mathbf{p}_{2})&=\sqrt{d_s^2+\left(\frac{d_c}{S}\right)^2m^2} \text{\ \ \ where\ }
\end{align}
\vspace{-0.5cm}
\begin{align*}
    d_s(\mathbf{p}_1,\mathbf{p}_{2})&=\|\mathbf{x}_1-\mathbf{x}_2\|_2, \text{ } \vspace{0.2cm}
    d_c(\mathbf{p}_1,\mathbf{p}_{2})=\|\mathbf{l}_1-\mathbf{l}_2\|_2,
\end{align*}

\noindent
where $D$ is Lab-space in the colour image case and just intensity in the grey-scale image case, and $m$ is a parameter which tunes the importance of spatial as compared to Lab-distance. The central observation exploited by SLIC is that if the spatial distance between two image points ($\|\mathbf{x}_1-\mathbf{x}_2\|_2$) is large, then the distance $d(\mathbf{p}_1,\mathbf{p}_{2})$ is large and its calculation can be spared in order to enhance computational efficiency.
While SLIC has demonstrated powerful results, it is more limited by its construction. These limitations are addressed and motivate our proposed solution.

\section{STRUCTURE ADJUSTING SUPERPIXELS} 

In this section, we address \emph{how the failure of SLIC is related to the modelling hypotheses, and how these failures motivate our proposed solution, dSLIC}.

Whilst SLIC provides fast and qualitative image segmentation,  one can observe - from the description given in the previous section - that SLIC is restricted by its own definition. Notably, these restrictions are two-fold. Firstly,  SLIC tends to segment large uniform regions in an image with more superpixels than are intuitively necessary. Secondly, in search domains with many boundaries, the resulting superpixels size is smaller than SLIC's search radius.


\textbf{Proposed Solution $-$ dSLIC.}~ The central idea to our approach is that the problems described above could be overcome by dynamically adjusting the search range according to the \emph{density of structure} in a given part of the image.

The presence of boundaries and hence of non-uniform parts in an image can be captured by the size of the discrete gradient of the image. Particularly, if we work with greyscale images with $D=[0,1]$ then $|DI|$ will be large on edges of the image and provide an indicator of such in the interval $[0,\sqrt{2}]$.  For colour images, we calculate the gradient and all following measures simply by converting to greyscale first. However, the gradient - as local descriptor of the image - does not allow determining if a given point belongs, or not, to an image part with many boundaries. Thus,  we propose the following measure of structure in the image:
\begin{align}
    f(x)=(g_\sigma\star |DI|)(x),
\end{align}

\noindent
where $g_\sigma$ is a Gaussian kernel of variance $\sigma^2$. In practise, we found that the following additional scaling is able to deal with any given input image:
\begin{align}
    f(x)=\frac{\left(g_{20}\star \left(|DI|\wedge\frac{2}{255}\right)\right)(x)}{\max_{x\in\mathcal{X}}\left(g_{20}\star \left(|DI|\wedge\frac{2}{255}\right)\right)(x)}.
\end{align}
%
%
\begin{figure}[t!]
\centering
\includegraphics[width=0.5\textwidth]{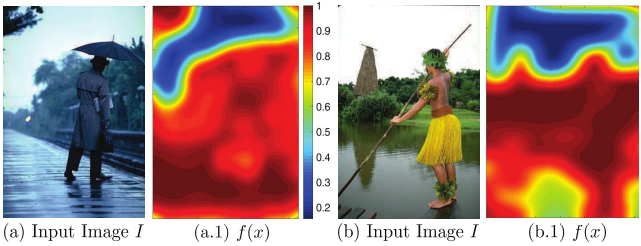}
\caption{Samples input (a) and (b) and the corresponding $f(x)$ plot. (a.1) and (b.1) illustrate the effect of $f(x)$, in which small values are related to uniform areas while high values to regions with rich structure.}
\label{fig::example_f}
\end{figure}
Our function $f(x)$ is plotted in Fig~\ref{fig::example_f}. From these examples, we can observe that $f(x)$ is small in uniform regions, such as the sky,  and large in non-uniform  regions such as the trees and people.
This comes from the fact that the normed gradient provides a local indicator for existent structure and the Gaussian convolution allows us to spread this information across a neighbourhood. In order to rescale the search radius appropriately we define:
\begin{align}
    g(x):=\exp\left(f(x)-\overline{f}\right),
\end{align}

\noindent
where $\overline{f}$ denotes the average of $f$ on the image grid, and propose the following dynamic distance computation:
\begin{center}
$ d((x,I(x)),F(S^{(t)}_i))\text{ if }|x-(F(S^{(t)}_i))_1|\leq 2Sg(F(S^{(t)}_i));$\\
$\text{Set }d((x,I(x)),F(S^{(t)}_i))=\infty\text{ o.w.}; $
\end{center}

The overall procedure of our dSLIC method - which dynamically adjusts the search field size according to our structure measure $g$ - is listed in Algorithm~\ref{alg::ours}.


\section{Experimental results}
In this section we describe in detail the experiments that we conducted to evaluate our dSLIC algorithm.

\textbf{Data Description.} We evaluated our dSLIC algorithm using images coming from three datasets: (i) The Berkeley Segmentation Dataset 500 ~\cite{arbelaez2011contour}, (ii) The Stanford Background Dataset~\cite{gould2009decomposing} and (iii) The Fashionista dataset~\cite{yamaguchi2012parsing,stutz2017superpixels}. All results presented in this section were run under the same condition, and using an Intel Core i7 CPU at 3.40 GHz-64GB.

\begin{algorithm}
 \begin{algorithmic}[1]
            \State Take an initial segmentation $\{S^{(0)}_i\}_{i=1}^k$\;
            \While{$E> $ threshold and $t<T$}
            \Procedure{Distance Calculations}{}
                \begin{align*}
                \begin{cases}
                \text{Compute }d((x,I(x)),F(S^{(t)}_i))\\\hspace{0.3cm}\text{ if }|x-(F(S^{(t)}_i))_1|\leq 2Sg(F(S^{(t)}_i));\\
                \text{Set }d((x,I(x)),F(S^{(t)}_i))=\infty\text{ o.w.};
                \end{cases}
            \end{align*}
            \EndProcedure

            \State Assign $x$ to the nearest cluster at time $t$ such that $S^{(t+1)}_i=\{x:d((x,I(x)),F(S^{(t)}_i))=\min_{1\leq j\leq n} d((x,I(x)),F(S^{(t)}_j))\}$\;

            \State Set $t\gets t+1$\;
            \State Compute residual error $E$.
            \EndWhile
       	\State \textbf{return} Final segmentation $\{S^{(t-1)}_i\}_{i=1}^k$
 \end{algorithmic}
 \caption{Structure Adjusting Superpixels (dSLIC)}
 \label{alg::ours}
\end{algorithm}
\begin{figure}[t]
\centering
\includegraphics[width=0.5\textwidth]{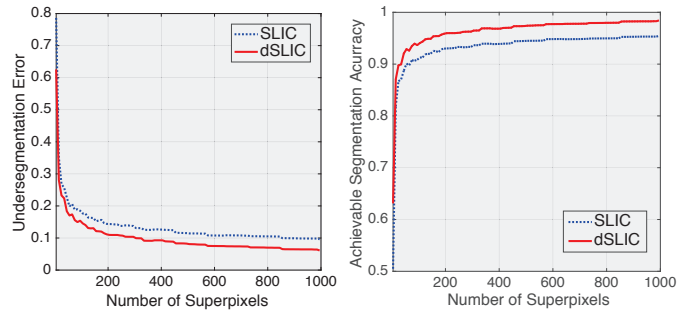}
\caption{Plots comparing dSLIC to SLIC using two performance evaluation metrics, undersegmentation error and segmentation accuracy.}
\label{fig:UnderSegB}
\end{figure}
\begin{figure*}[t]
\centering
\includegraphics[width=0.97\textwidth]{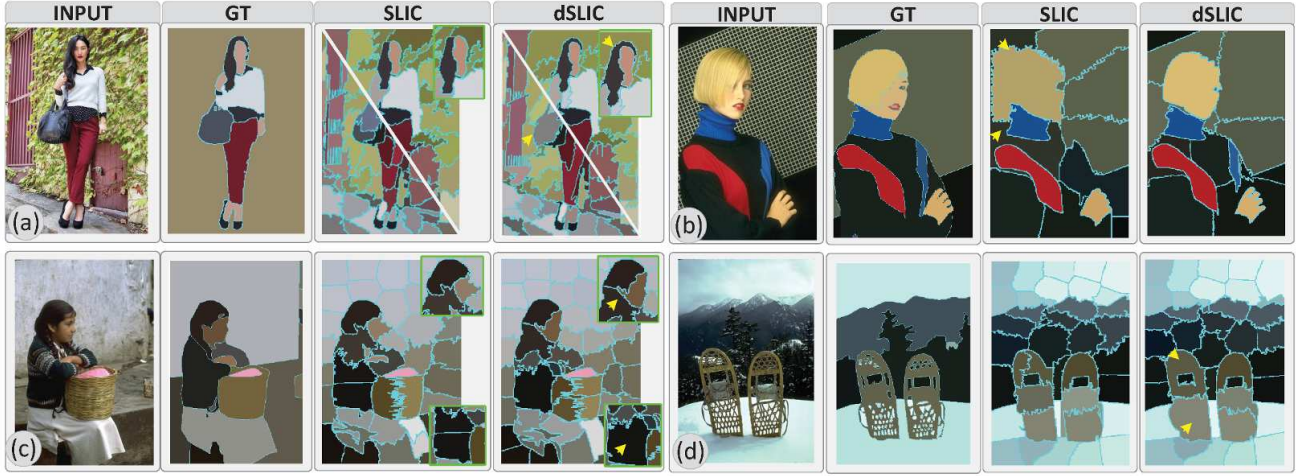} 
\caption{Superpixel segmentation examples. From left to right: input image, ground truth (GT), SLIC and our dSLIC outputs from four sample images. Visual assessment shows that the proposed algorithm performs better than SLIC as it respects object boundaries and tends to divide an image into uniform regions (see yellow arrows).}
\label{fig::slicvsdslic}
\end{figure*}
\begin{figure}[h!]
\centering
\includegraphics[width=0.5\textwidth]{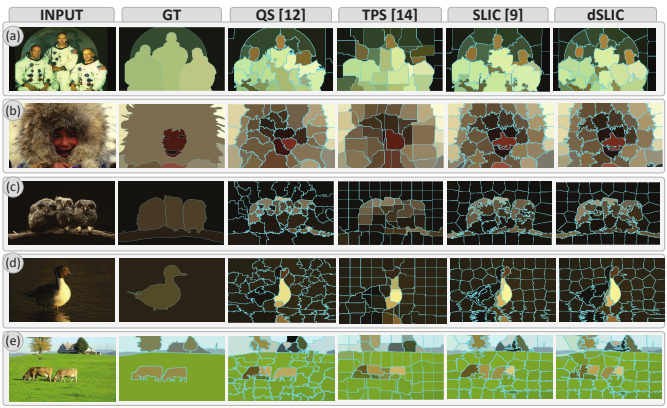}
\caption{Superpixel segmentation examples. From left to right: input image, ground truth and three approaches from
the body of literature~\cite{SLIC::2012,vedaldi2008quick,tang2012topology}. See text for discussion.}
\label{fig:res1A}
\end{figure}
\textbf{Results.} We divided our evaluation scheme in two parts: (1) The first, which is\emph{ the main focus }of this work, demonstrates the advantages of our solution (dSLIC) - from both qualitative and quantitative points of view -  against the SLIC algorithm. (2) We offer an initial insight comparison between dSLIC and some works from the state-of-the-art.
In this work, in (1) we set $m=20$ since it yields balance between uniformity of shape and boundary adherence.

We started with visual evaluation of our approach compared against SLIC. Upon visual inspection of the results in Fig.~\ref{fig::slicvsdslic}, we can observe that  dSLIC  is able to combat the problems encountered with SLIC, in that it samples larger superpixels into uniform regions, avoiding in this way segmenting the input images into unnecessarily small superpixels. More precisely, this positive effect is illustrated in the sample output in Fig.~\ref{fig::slicvsdslic}, in which we highlight from (a) both the hair and pants that were more properly grouped into the same superpixels. SLIC in that case failed to fully segment the hair and added avoidable segments to the pants. dSLIC was also more successful on properly identifying the face and the background in (b) where those details were lost in the results from SLIC. Similarly, the segmentation results of the woman in (c) show proper grouping of the face and the sweater.
%
%

To further evaluate the results, we also offer quantitative analysis based on two well-used metrics for superpixels evaluation. The first metric is undersegmentation error which measure conformity to the true boundaries.  The second metric is the achievable segmentation accuracy (ASA) which is a performance upperbound measure that gives higher achievable accuracy when superpixels are utilised as units for object segmentation. These performance metrics are plotted against the number of superpixels for both SLIC and dSLIC approaches, which were obtained from segmented 15\% of randomly selected images of \textit{all datasets}.

The plot on the left side of Fig.~\ref{fig:UnderSegB} shows the undersegmentation error curves of both approaches where we can observe that the proposed algorithm outperforms SLIC at all the superpixel counts and reduces the error rate by more than 20\%. Similarly, the ASA curves - that appear on the right side of Fig.~\ref{fig:UnderSegB} - show that dSLIC yields a better achievable segmentation upperbound at all the superpixel counts. With dSLIC, the ASA is 95\% with 200 superpixels where the same accuracy can only be achieved with 1000 superpixels for SLIC. This  improvement comes at a negligible cost in runtime ($\sim 2\%$).

While the main aim of this work is to offer evidence about the performance improvement of our approach over SLIC, we also offer an insight comparison of our solution against some work from the state of the art, particularly \cite{vedaldi2008quick,tang2012topology}. All results were adapted to the same conditions for fair comparison. Fig.~\ref{fig:res1A} shows selected images comparing the results of each of the compared approaches. Visual assessment of the figure shows that our solution was able to identify shapes' details and merge regions with uniform domain. Clear examples can be seen in the segmentation of the faces and the moon structure in (a) and the face and wooden branch in (b) and (c) respectively. Our algorithm was also able to properly unify small details in the image such as the duck's beak and the house structure in examples (d) and (e). We also noticed that among the compared algorithms, SLIC and dSLIC demanded the lowest computational load.

\section{Conclusions}
We have considered the problem of superpixel image segmentation, which is of theoretical interest and practical importance. We build on the state-of-the-art algorithm SLIC \cite{SLIC::2012} by introducing a dynamic search range based on a structure measure. This helps avoid both unnecessary oversegmentation of uniform areas, and repeatedly searching object-dense areas. Numerical experiments showed evidence that our proposed algorithm, dSLIC, outperforms the SLIC solution in terms of undersegmentation error (20\%) and achievable accuracy. Visual assessment of the results confirms that dSLIC provides more meaningful superpixels in the same linear runtime.
\textit{This work offers an initial proof of concept of our method dSLIC.} Future work may include an extensive comparison and analysis against more works from the state-of-the-art.


\bibliographystyle{IEEEbib}

\begin{thebibliography}{99}

\bibitem{ngo2015automatic}
Ngo, Tran-Thanh and Collet, Christophe and Mazet, Vincent.
Automatic rectangular building detection from VHR aerial imagery using shadow and image segmentation.
IEEE International Conference on Image Processing (ICIP).
pp. 1483--1487, 2015.

\bibitem{huang2016stem}
Huang, Xinyu and Li, Chen and Shen, Minmin and Shirahama, Kimiaki and Nyffeler, Johanna and Leist, Marcel and Grzegorzek, Marcin and Deussen, Oliver.
Stem cell microscopic image segmentation using supervised normalized cuts.
IEEE International Conference on Image Processing (ICIP),
pp. 4140--4144, 2016.

\bibitem{shen2017efficient}
Shen, Haocheng and Zhang, Jianguo and Zheng, Weishi.
Efficient symmetry-driven fully convolutional network for multimodal brain tumor segmentation.
IEEE International Conference on Image Processing (ICIP), 2017.

\bibitem{yan2015object}
Yan, Junjie and Yu, Yinan and Zhu, Xiangyu and Lei, Zhen and Li, Stan Z.
Object detection by labeling superpixels.
IEEE Conference on Computer Vision and Pattern Recognition (CVPR), pp. 5107--5116, 2015.

\bibitem{liu2015deep}
Liu, Fayao and Shen, Chunhua and Lin, Guosheng.
Deep convolutional neural fields for depth estimation from a single image.
IEEE Conference on Computer Vision and Pattern Recognition (CVPR),
pp. 5162--5170, 2015.

\bibitem{lu2013patch}
Lu, Jiangbo and Yang, Hongsheng and Min, Dongbo and Do, Minh N.
Patch match filter: Efficient edge-aware filtering meets randomized search for fast correspondence field estimation.
IEEE Conference on Computer Vision and Pattern Recognition (CVPR), pp. 1854--1861, 2013.

\bibitem{menze2015object}
Menze, Moritz and Geiger, Andreas.
Object scene flow for autonomous vehicles.
IEEE Conference on Computer Vision and Pattern Recognition (CVPR), pp.3061--3070, 2015.


\bibitem{wang2011superpixel}
Wang, Shu and Lu, Huchuan and Yang, Fan and Yang, Ming-Hsuan.
Superpixel tracking.
IEEE International Conference on Computer Vision (ICCV), pp. 1323--1330, 2015.


\bibitem{SLIC::2012}
Achanta, Radhakrishna and Shaji, Appu and Smith, Kevin and Lucchi, Aurelien and Fua, Pascal and S{\"u}sstrunk, Sabine.
SLIC superpixels compared to state-of-the-art superpixel methods.
IEEE Transactions on Pattern Analysis and Machine Intelligence (PAMI),
vol.34, pp. 2274--2282, 2012.


\bibitem{stutz2017superpixels}
Stutz, David and Hermans, Alexander and Leibe, Bastian.
Superpixels: An evaluation of the state-of-the-art.
Computer Vision and Image Understanding, 2017.

\bibitem{ren2003learning}
Ren, Xiaofeng and Malik, Jitendra.
Learning a classification model for segmentation.
IEEE International Conference on Computer Vision (ICCV),
2003.

\bibitem{vedaldi2008quick}
Vedaldi, Andrea and Soatto, Stefano.
Quick shift and kernel methods for mode seeking.
European Conference on Computer Vision (ECCV),
pp. 705--718, 2008.

\bibitem{van2012seeds}
Van den Bergh, Michael and Boix, Xavier and Roig, Gemma and de Capitani, Benjamin and Van Gool, Luc.
SEEDS: Superpixels extracted via energy-driven sampling.
European Conference on Computer Vision (ECCV), pp. 13--26,
2012.

\bibitem{tang2012topology}
Tang, Dai and Fu, Huazhu and Cao, Xiaochun.
Topology preserved regular superpixel
IEEE International Conference on Multimedia and Expo (ICME),
pp. 765--768, 2012.

\bibitem{lloyd1982least}
Lloyd, Stuart. Least squares quantization in PCM.
IEEE transactions on information theory, vol. 28, pp. 129--137, 1982.


\bibitem{arbelaez2011contour}
Arbelaez, Pablo and Maire, Michael and Fowlkes, Charless and Malik, Jitendra.
Contour detection and hierarchical image segmentation.
IEEE Transactions on Pattern Analysis and Machine Intelligence (PAMI),
vol. 33, pp. 898--916, 2011.


\bibitem{gould2009decomposing}
Gould, Stephen and Fulton, Richard and Koller, Daphne.
Decomposing a scene into geometric and semantically consistent regions.
IEEE International Conference on Computer Vision (ICCV),
pp. 1--8, 2009.



\bibitem{yamaguchi2012parsing}
Yamaguchi, Kota and Kiapour, M Hadi and Ortiz, Luis E and Berg, Tamara L.
Parsing clothing in fashion photographs.
IEEE Conference on Computer Vision and Pattern Recognition (CVPR),
 pp. 3570--3577, 2012.


%


\end{thebibliography}

\end{document}